\documentclass[sigconf]{acmart}

\usepackage{float}
\usepackage{placeins}
\usepackage{algorithm}
\usepackage{algorithmicx}
\usepackage{algpseudocode}
\usepackage{colortbl}
\usepackage{fixltx2e}
\usepackage{dblfloatfix}
\usepackage{url}
\theoremstyle{definition}
\newtheorem{hypothesis}{Hypothesis}
\setcopyright{rightsretained}%\setcopyright{usgov}%\setcopyright{usgovmixed}%\setcopyright{cagov}%\setcopyright{cagovmixed}%\setcopyright{none}%\setcopyright{acmcopyright}%\setcopyright{acmlicensed}

\acmDOI{XX.XXX/XXX_X}
\acmISBN{XXX-XXXX-XX-XXX/XX/XX}
\acmConference[GECCO '17]{the Genetic and Evolutionary Computation Conference 2017}{July 15--19, 2017}{Berlin, Germany}
\acmYear{2017}
\copyrightyear{2017}
\acmPrice{15.00}

\begin{document}
\title{Hierarchical Surrogate Modeling for Illumination Algorithms}
\author{Alexander Hagg}

\affiliation{%
  \institution{Bonn-Rhein-Sieg University of Applied Sciences}
  \streetaddress{Grantham-Allee 20}
  \city{Bonn} 
  \state{Germany} 
  \postcode{53757}
}
\email{alexander.hagg@h-brs.de}

\begin{abstract}
Evolutionary illumination is a recent technique that allows producing many diverse, optimal solutions in a map of manually defined features. To support the large amount of objective function evaluations, surrogate model assistance was recently introduced~\cite{gaier2017feature}. Illumination models need to represent many more, diverse optimal regions than classical surrogate models.
In this PhD thesis, we propose to decompose the sample set, decreasing model complexity, by hierarchically segmenting the training set according to their coordinates in feature space. An ensemble of diverse models can then be trained to serve as a surrogate to illumination.
\end{abstract}


\maketitle

\section{Introduction}
\label{sec:intro}

Computer-automated design was first introduced in 1963 by Kamentsky and Liu \cite{kamentsky1963computer}, who created a computer program to design character recognition logic based on the processing of data samples. Many applications and methods have since been developed. Today, optimization of high-dimensional problems, such as in computational fluid dynamics~\cite{Emmerich2004} or robotics~\cite{Cully2015}, can be extremely expensive in terms of computational effort. 

To decrease the necessary effort to optimize an expensive objective function, approximative models are used to serve as a \textit{surrogate} for these simulations. \textit{Surrogate-assisted optimization} (SAO) is a technique which, supported by models from machine or statistical learning, is typically used when the objective function is complex, data is scarce or evaluation of potential solutions is expensive. 
A surrogate model only needs to learn an approximation of the objective function, also called the \textit{response surface}, close to optimal solutions. In evolutionary optimization (EO) specifically, the approximation accuracy requirement devolves to a ranking accuracy requirement. The model only needs to accurately compare solutions. 

The necessary number of evaluations can be further reduced by using Bayesian optimization (BO) \cite{Brochu2010}. The technique uses a prior over the objective function and evidence from known samples to select the best next observation based on a utility function, also called an \textit{acquisition function}. This function balances exploration, sampling from uncertain areas, and exploitation, choosing samples that are most likely to perform well.

Both techniques, surrogate-assistance and the acquisition function, are used for \textit{data efficient learning}. In this paradigm, learning requires a modeling technique that can not only accurately predict the objective function, but also estimate its prediction confidence. 

SAO is used to find a single solution, or a Pareto front of solutions in multi-objective optimization. To find more interesting designs or design principles, \textit{illumination} algorithms, introduced by Mouret et al.~\cite{Mouret2015}, are used to explore the relationship between user-defined features and the maximal performance. For example, if an engineer wants to design a car, the relationship between the volume of the car's luggage space and its turning radius can be illuminated. The engineer can use this feature relationship as a basis for design decisions. The method needs many evaluations of the objective function, which can be greatly reduced using surrogate-assisted illumination (SAIL), which was introduced by Gaier et al.~\cite{gaier2017feature}.

\section{Surrogate-Assisted Optimization}
\begin{figure}[h!tbp]
	\includegraphics[width=0.6\linewidth]{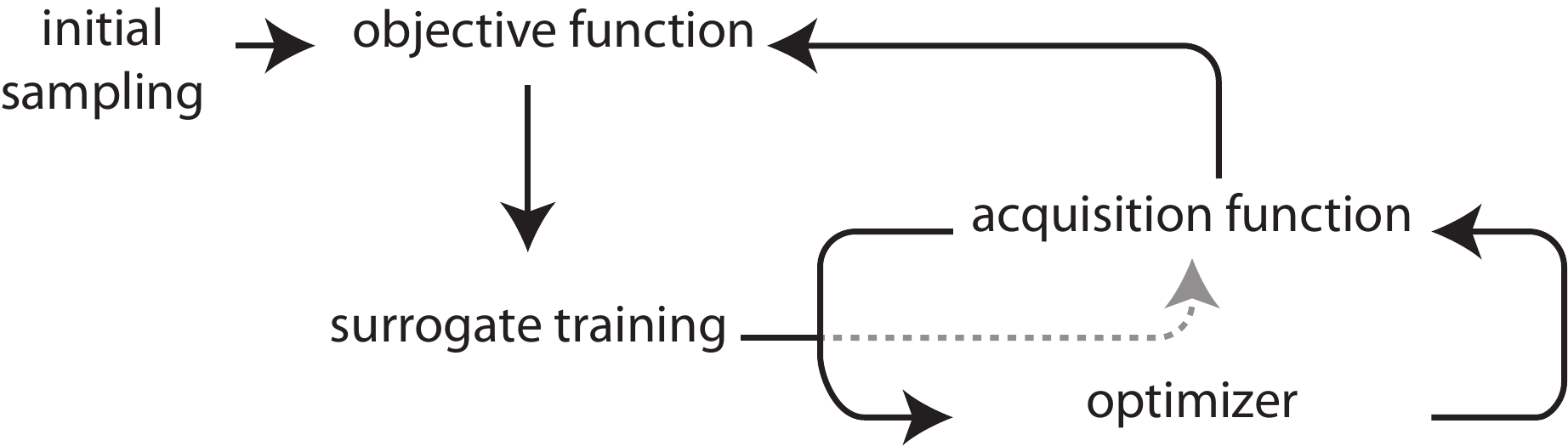}
	\caption{Surrogate assisted optimization.}
	\label{img:SAO}
\end{figure}

The SAO process is depicted in Figure \ref{img:SAO}. Initial training of the surrogate model is performed on a small training set, which is evenly spread over parameter space.
The expensive objective function is used to retrieve samples' fitness value. Then, a surrogate model is created based on the initial sample set.

The surrogate model is updated in an on-line learning strategy. The optimization strategy is used to \textit{explore} the model to find optimal solutions. In Bayesian optimization this is not done directly. To acquire new samples, the model's confidence about its prediction is included. A common acquisition function is upper confidence bound (UCB) sampling~\cite{Auer2003}. Here, the uncertainty of the model is added to the predicted fitness to overestimate the solution's objective performance. The weighting of fitness and confidence controls the exploratory and exploitative behavior of the optimizer. This helps lowering \textit{regret}: even if the sample is not of high fitness, the model will benefit from this additional knowledge.

Because Gaussian process (GP) models are effective with small sample sets and include an uncertainty measure, they are often used in optimization~\cite{Emmerich2004,rasmussen2006gaussian,Brochu2010,Jin2011,Cully2015,gaier2017feature}. A GP describes a random distribution of functions which are defined by the mean function \textit{m} and the covariance function \textit{k}. GP models are interpolative. The predicted value depends on the proximity to a known sample and its value, which is defined by the covariance function.

\begin{figure}[htbp]
	\includegraphics[width=1\linewidth]{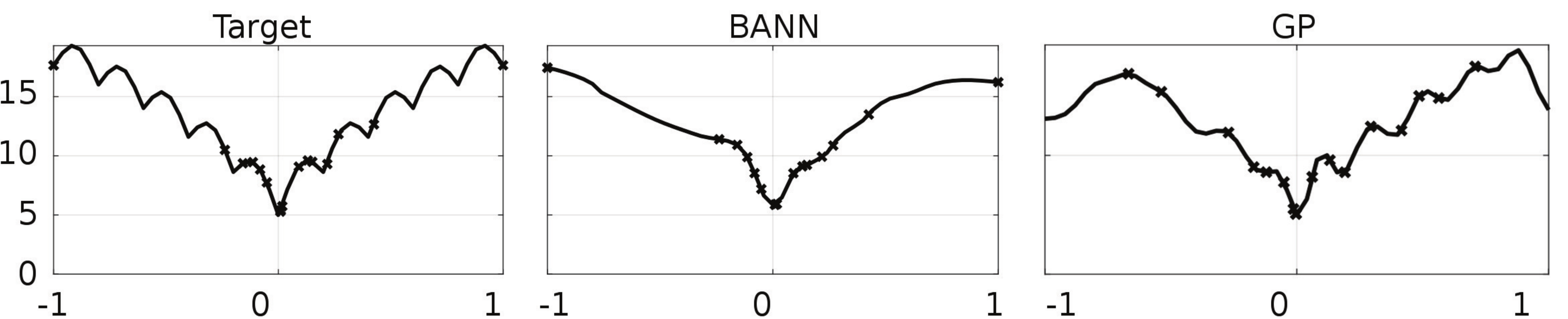}
	\caption{It is easier to find the global minimum of the objective function (left) in the smoother surrogate model (middle) than in the more accurate one (right).}
	\label{img:smoothness}
\end{figure}

Optimization performance does not depend on the surrogate model's prediction accuracy alone. Depending on the chosen optimization technique, \textit{overfitting} of the surrogate model can have an adverse effect on the convergence 
speed of the optimizer. As can be seen in Figure \ref{img:smoothness}, in gradient based approaches, a model which does not fit the data perfectly can 
offer a smoother, easier traversable response surface, a phenomenon also called the \textit{blessing of uncertainty}, which is described in work by Ong et al.~\cite{Yew-SoonOng2006}. 
The same is valid in EO but to a lesser degree. Even though EO only depends on rank comparisons between solutions, they still depend on a \textit{virtual gradient}, which is induced by the fact that solutions are only mutated in small steps. Local optima can still lead the algorithm into \textit{traps}, leading to longer convergence times~\cite{Yew-SoonOng2006}.

EO have shown to be better at divergent search than classical, gradient-based approaches~\cite{Crepinsek2013}. Whereas \textit{exploitation} of existing knowledge to improve solutions is an aspect that is central to both fields, many EO techniques exist that focus on \textit{exploration} of the solution space, also known as \textit{novelty search}. All techniques benefit from surrogate models that are not too accurate, creating a smoother objective function for the optimizer.

\section{Surrogate Assisted Illumination}
\label{sec:SAIL}

\begin{figure}[htbp]
	\includegraphics[width=1\linewidth]{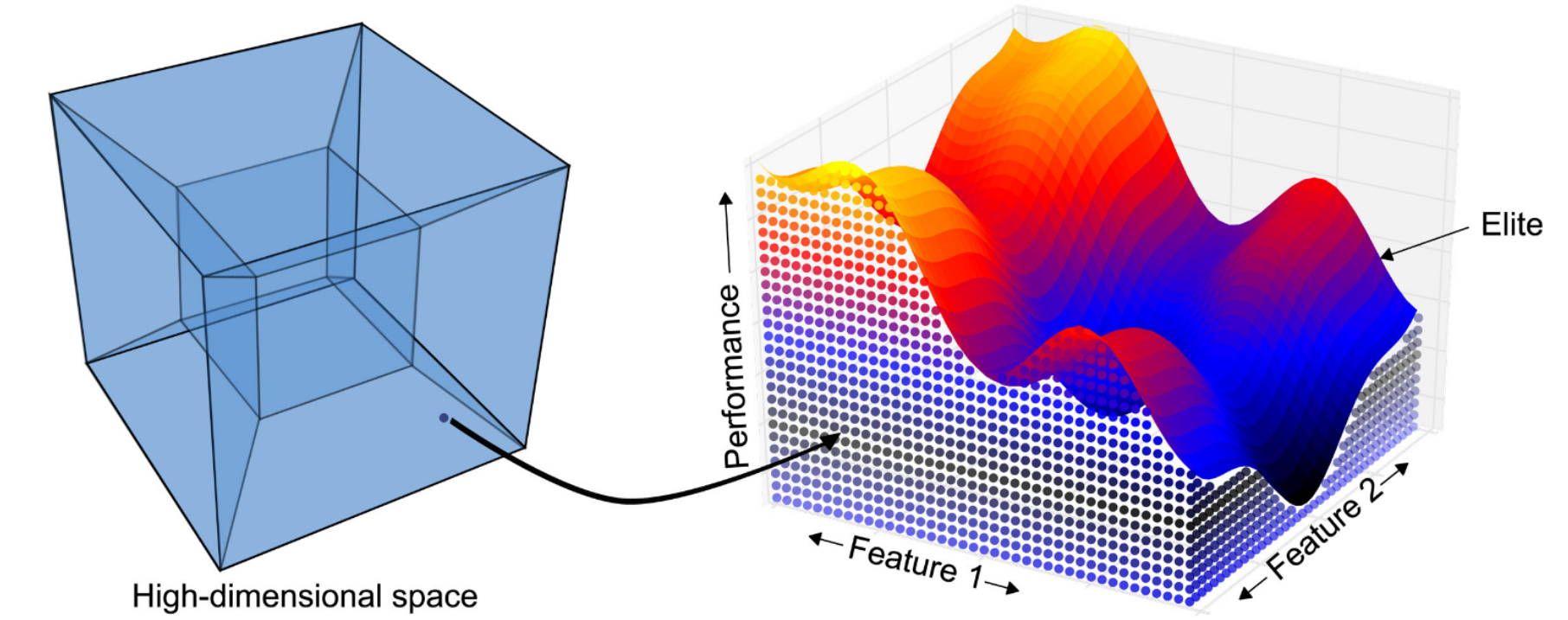}
	\caption{MAP-Elites projects high-dimensional samples onto a low-dimensional feature map~\cite{Mouret2015}. The surface on the right shows the elites' objective function values.}
	\label{img:mapelites}
\end{figure}

In recent work, Mouret and Clune introduced~\cite{Mouret2015} a new divergent optimization technique, MAP-Elites, which can illuminate the relationship between features and their impact on optimal solutions. The technique is a mixture of optimization and novelty search. Solutions are mapped from their high-dimensional parameter space onto a lower-dimensional map of features (Figure \ref{img:mapelites}). The space is discretized into bins, niches in which individuals are similar with respect to the selected features. 

To initialize MAP-Elites, a set of random solutions is first evaluated and assigned to bins. If a bin is empty, the solution is placed inside. If another solution is already occupying the bin, the new solution replaces it if it has a higher fitness, otherwise it is discarded. As a result, each bin contains the best solution found so far (\textit{elite}). 
To produce new solutions, parents are chosen randomly from the elites, then mutated and evaluated, and finally assigned to a bin based on their feature values.
Child solutions have two ways of joining the breeding pool, either by discovering an unoccupied bin or out-competing an existing solution for its bin. By repeating this process, the feature space gets increasingly explored, resulting in an increasingly optimal collection of solutions.

The technique allows engineers to generate a large number of optimal solutions that can be used to easily switch strategies in robotics control~\cite{Cully2015} or to model the optimal solutions in a feature space~\cite{gaier2017feature}. 

SAIL is an extension to this algorithm. Gaier et al.~\cite{gaier2017feature} use a GP model to support the illumination process by UCB sampling to add knowledge to the surrogate model during optimization. The GP model needs to predict the objective function based on all 11 dimensions of the parameter space. In the evaluation case that was defined, optimization of a 2D airfoil, the number of dimensions and samples is not very large. When optimizing 3D shapes however, the dimensionality can easily go up to 88 or 96 dimensions~\cite{hasenjager2005three,brezillon2009aerodynamic}.

\section{Surrogate Methods and their Deficits}
In many aspects, SAIL is very similar to SAO. Both are online methods that require a retrainable model. They can both benefit from models that annotate their prediction with a confidence interval. In both cases, the amount of samples needed for an accurate model should be minimal, as evaluations are usually expensive. Finally, they both benefit from smooth models. 

On top of this, SAIL needs millions of comparisons, depending on the map's resolution, many more than most SAO methods. In SAO, the training of the surrogate is often more expensive than evaluation, but since SAIL needs to evaluate the model more often, models with low evaluation times are beneficial to the computational requirements of the optimization process. 

\begin{figure}[h!tbp]
	\includegraphics[width=1\linewidth]{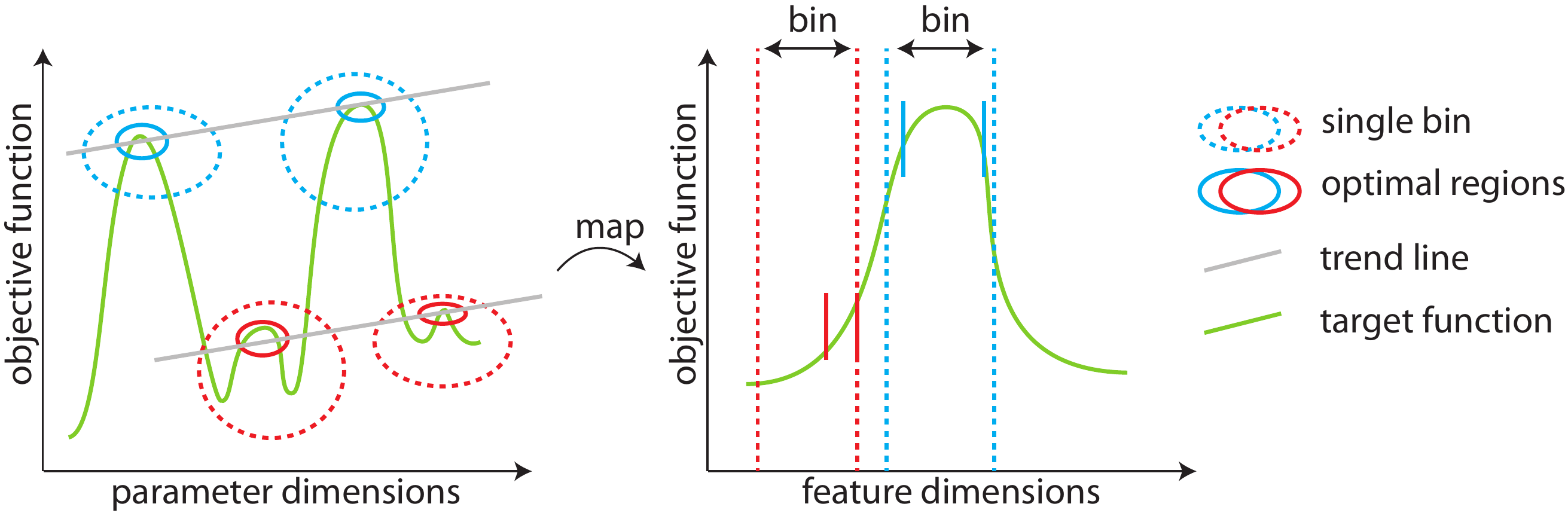}
	\caption{Left: parameter space and objective function for surrogate model. Right: map produced by non-linear feature transformation.}
	\label{img:SAIL:target}
\end{figure}

Another difference is the fact that SAIL is divergent and finds many, \textit{diverse} optima. Locally accurate surrogate models could be too expensive to accurately model for example the optimal regions for 64x64 bins. In this case, 4096 local models would have to be trained. Also, since the diversity of the training set is high and bins are defined by non-linear features, the optimal regions within the bins do not necessarily belong to a continuous optimal region in the objective function (Figure \ref{img:SAIL:target}). 
In this simplified example, notice that samples belonging to the same bin can belong to different regions in parameter space. Within each bin, the surrogate model only needs to be accurate towards the more optimal solutions, but this region can consist of solutions from multiple regions in parameter space. A single surrogate model would have to learn the high-order polynomial (green), although for rank comparisons in illumination it can be sufficient to learn linear trend lines (gray).

\begin{figure}[htbp]
	\includegraphics[width=0.6\linewidth]{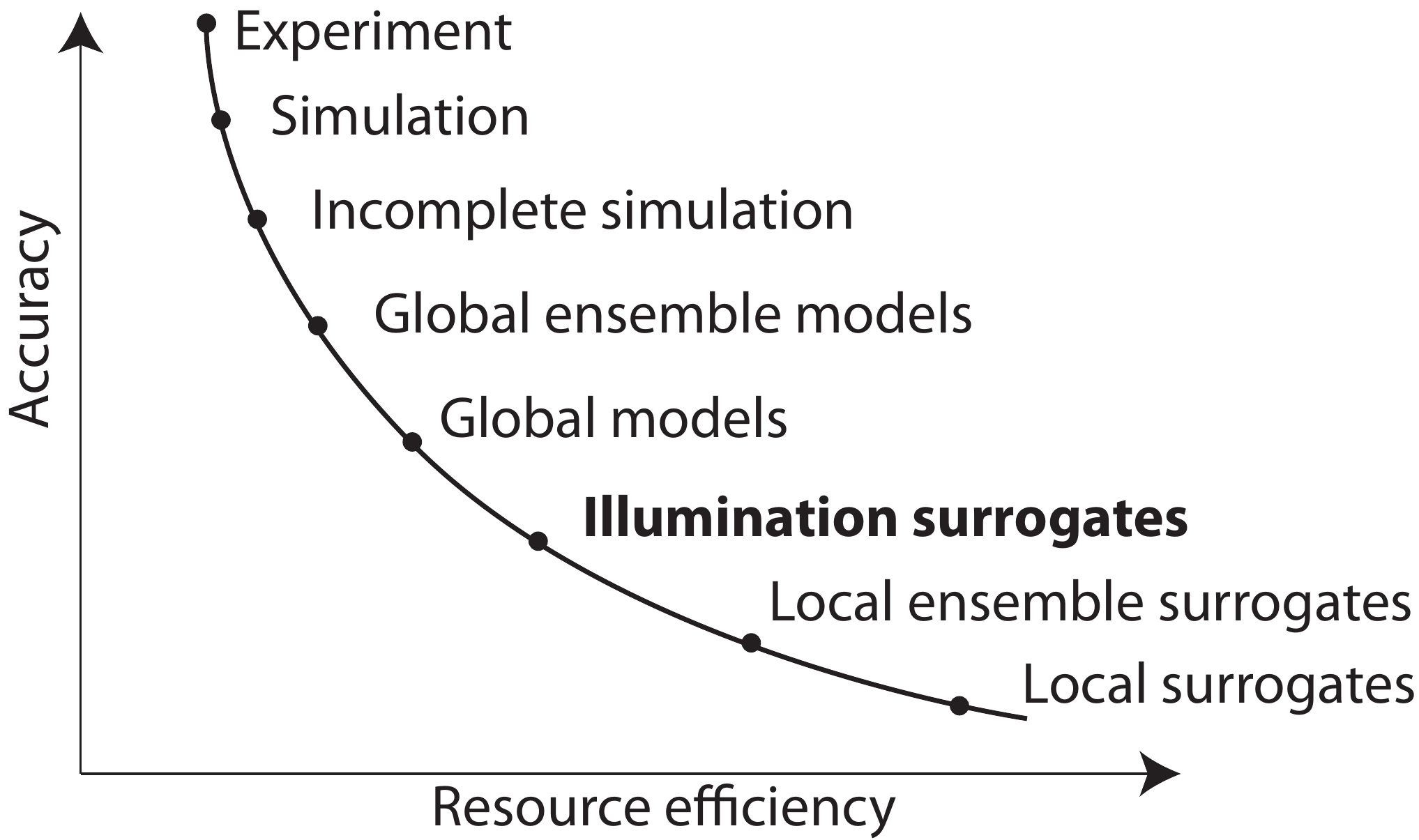}
	\caption{Trade-off between accuracy and resource efficiency~\cite{Jin2011}.}
	\label{img:models}
\end{figure}

Training a global model that approximates the entire objective function is not feasible in optimization, as samples are expensive and we will not be able to create sufficient samples to train such a high-dimensional model. A contradiction seems to have opened up. On one hand, in order to have models that allow comparing all individuals in a bin, a local model could be trained that specializes on this task. On the other hand, tractability will be lost if every bin gets its own local model. We therefore need to train models that are more general than local surrogates but less complex than global models, placing illumination surrogates between the two techniques in terms of accuracy and efficiency (Figure \ref{img:models}).

SAIL will push GP to its limits in terms of computational complexity when applied to real world problems. Training complexity is $\mathcal{O}(n^3)$ ($n =$ number of samples) for training and $\mathcal{O}(n^2)$ for prediction~\cite{Quinonero}. Optimized versions exist, reducing the complexity to $\mathcal{O}(m^2n)$ for training and $\mathcal{O}(m^2)$ for prediction by reducing the number of interpolation points using $m$ pseudo-inputs~\cite{Snelson2006}. 

Ensemble methods, like bootstrapped artificial neural networks (BANN)~\cite{Breiman1996}, can provide confidence measures as well, by training a \textit{homogeneous} set of models. The ensemble's prediction is calculated by taking the mean of all its members. A confidence measure is obtained by evaluating the variation in member predictions.

\begin{figure}[htbp]
	\includegraphics[width=0.8\linewidth]{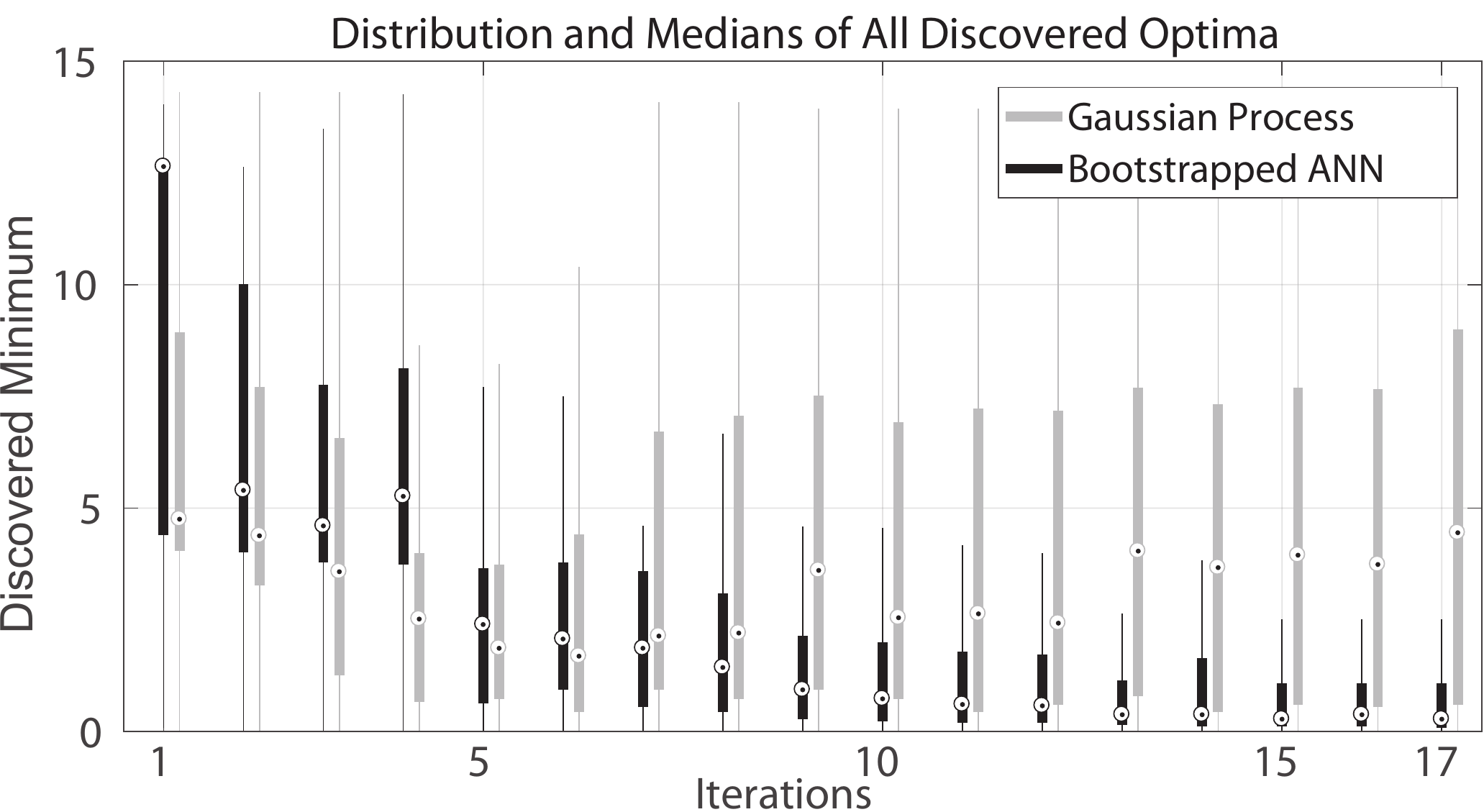}
	\caption{Comparison of surrogates assisting hill climber.}
	\label{img:gpvsbann}
\end{figure}

In a preliminary investigation, we compared GP with a BANN (trained with Levenberg-Marquardt~\cite{levenberg1944method}) using a hill climber to optimize a 1D Ackley function.
The hill climber was initialized from 10 fixed equidistant starting locations in every run in order to show how a naive optimization algorithm would perform from any starting position. The experiment was replicated 100 times. Figure \ref{img:gpvsbann} shows all optima that were found. Although the GP-assisted runs converge quicker than the BANN, most runs do not reach the global optimum. The median and variance of the discovered minima is much larger than the ones found using BANN-assisted hill climbing. This is due to the BANN model being much smoother, as was shown in Figure \ref{img:smoothness}. GP-assisted illumination might not benefit as much from the blessing of uncertainty as do more global models like artificial neural networks (ANN).

\section{Proposal}
\label{sec:prop}

In order to support illuminating a high-dimensional search space, we propose to segment the objective function into simpler functions, along the same feature dimensions used by the illumination algorithm. This decomposition can simplify the modeling process, by allowing the optimal region in a bin, which can be disparate in parameter space, to be approximated by lower complexity models. 

\begin{figure}[htbp]
	\includegraphics[width=1\linewidth]{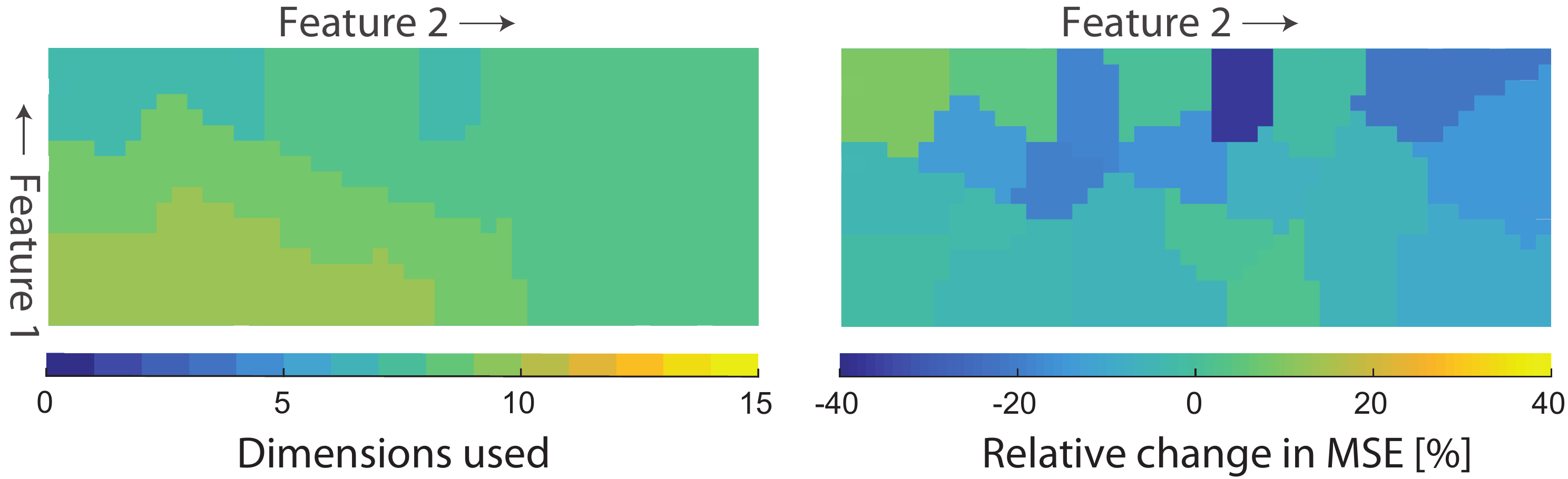}
	\caption{Dimensionality reduction and changes in modeling errors on airfoil optimization map.}
	\label{img:dimredux}
\end{figure}

Figure \ref{img:dimredux} shows a preliminary result on modeling 2D airfoils, produced by MAP-Elites. The airfoils are defined by 15 parameters, typical airfoil features are used for the map, and the objective function is the aerodynamic drag coefficient of the airfoils. The sample set is segmented using unbiased k-means. The dimensionality in every segment is reduced using principal component analysis and dimensions which explain less than 1\% of total variation removed. ANN surrogates are trained to predict the drag coefficient.
The segmentation leads to a varying dimensionality reduction (left) in all segments. In many segments, the error, compared to the error of a flat model trained on all samples, is reduced as well.

\begin{hypothesis}
	Using segmentation of training samples in feature space, decomposing the problem of modeling the objective function can lead to dimensionality reduction, simplifying training. 
\end{hypothesis}

\begin{figure}[h!]\centering
	\includegraphics[width=0.6\linewidth]{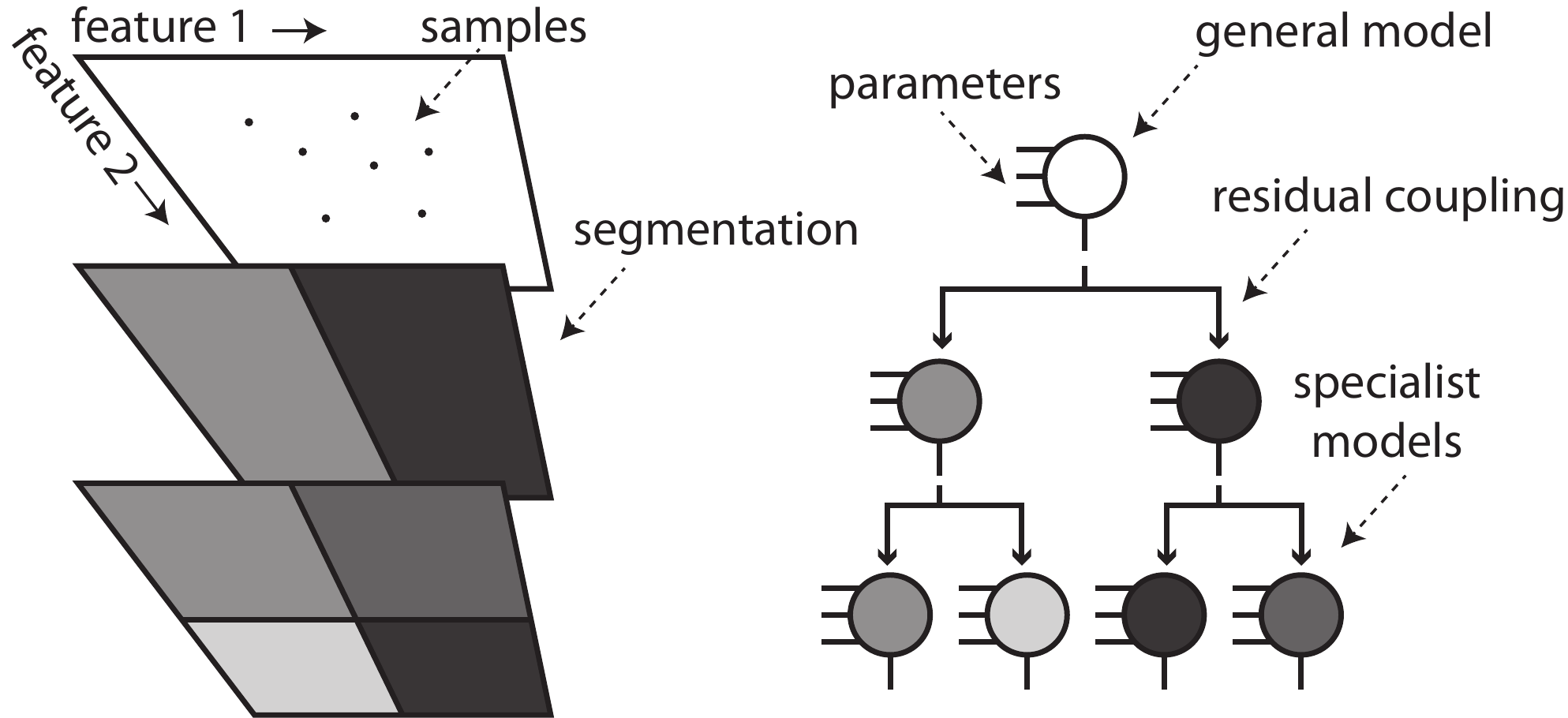}
	\caption{ Hierarchical decomposition of feature map with accompanying hierarchical surrogate model.}
	\label{fig:approach-abstract}
\end{figure}

Using separate surrogates for all training set segments might be too naive, as some bins could be described by the same model and some bins might need complex models that cannot be trained with only a small amount of samples. By using a hierarchical decomposition, we can create more general models, those that are trained on large, continuous regions of the fitness space, as well as more specialized models, reusing samples in many ways. This idea is borrowed from deep learning and other hierarchical approaches in computer vision~\cite{Behnke2003}.
Figure \ref{fig:approach-abstract} shows how a hierarchical decomposition of the map could be directly mapped onto a surrogate model structure. In the example above (Figure \ref{img:dimredux}), every submodel on layer $n$ is connected to their parent models using residual coupling~\cite{He2015}: models are trained to predict the discrepancy between the parent model and samples' true values.

An open question is whether to use a direct coupling between models, or to separately train models on various-sized sample subsets. 
In SAO and SAIL, training models is relatively cheap, because we do not have many samples. We are also willing to invest some time to reduce the amount of necessary real fitness evaluations. 
By training models on many subsets of the training data, we can maximize model diversity. We can use these diverse models to create a (hierarchical) ensemble, similar to \textit{bagging}, whereby multiple models are trained to decrease the total prediction variance~\cite{Breiman1996}.
Gu~\cite{Gu2016} showed that diverse, \textit{heterogeneous} ensembles often show a significantly higher accuracy than the base classifiers on their own.

\begin{hypothesis}
	Hierarchical decomposition of training samples can be used to train a diverse ensemble of surrogate models, allowing us to learn complex structure by using many shallow models.
\end{hypothesis}

The large set of diverse models can be used to extract confidence intervals, but it is unclear how to combine the predictions from ensemble members in a hierarchy. Whereas in normal ensemble learning, we can straightforwardly look at the prediction variance over all members to construct confidence intervals, this might not be the best option for hierarchical ensembles. \textit{Specialist} members are expected to be more accurate than the more \textit{general} models. We can therefore make a more informed weighting of models' prediction, putting different confidence levels to models in different layers of the hierarchy.

\begin{hypothesis}
	Confidence intervals can be determined from a model hierarchy in a more informed way than by direct variance estimation.
\end{hypothesis}

The working hypotheses do not include a specific modeling technique. In order to evaluate them, we will look at three main techniques: backpropagation training (Levenberg-Marquardt), GP models and neuroevolution. In the latter case, by training a diverse, large number of ANNs on hierarchical training segments, we mimic the mini-batch training method that was introduced by Morse et al.~\cite{Morse2016}. This technique showed that neuroevolution can rival stochastic gradient descent training on large training problems.

We aim to increase the surrogate's accuracy and the diversity of illuminated solutions in a data efficient learning context by using hierarchical surrogate models. 
By combining hierarchical surrogate modeling and ensemble techniques, we reuse samples in multiple ways, increasing the probability that an accurate surrogate is found.
We further reduce illumination convergence time by using smoother surrogate models. Finally, we hope to contribute a data efficient model which can be used in SAO, as well as in SAIL.

\bibliographystyle{ACM-Reference-Format}
\bibliography{student-workshop-2017} 

%%% -*-BibTeX-*-
%%% Do NOT edit. File created by BibTeX with style
%%% ACM-Reference-Format-Journals [18-Jan-2012].

\begin{thebibliography}{00}

%%% ====================================================================
%%% NOTE TO THE USER: you can override these defaults by providing
%%% customized versions of any of these macros before the \bibliography
%%% command.  Each of them MUST provide its own final punctuation,
%%% except for \shownote{}, \showDOI{}, and \showURL{}.  The latter two
%%% do not use final punctuation, in order to avoid confusing it with
%%% the Web address.
%%%
%%% To suppress output of a particular field, define its macro to expand
%%% to an empty string, or better, \unskip, like this:
%%%
%%% \newcommand{\showDOI}[1]{\unskip}   % LaTeX syntax
%%%
%%% \def \showDOI #1{\unskip}           % plain TeX syntax
%%%
%%% ====================================================================

\ifx \showCODEN    \undefined \def \showCODEN     #1{\unskip}     \fi
\ifx \showDOI      \undefined \def \showDOI       #1{{\tt DOI:}\penalty0{#1}\ }
  \fi
\ifx \showISBNx    \undefined \def \showISBNx     #1{\unskip}     \fi
\ifx \showISBNxiii \undefined \def \showISBNxiii  #1{\unskip}     \fi
\ifx \showISSN     \undefined \def \showISSN      #1{\unskip}     \fi
\ifx \showLCCN     \undefined \def \showLCCN      #1{\unskip}     \fi
\ifx \shownote     \undefined \def \shownote      #1{#1}          \fi
\ifx \showarticletitle \undefined \def \showarticletitle #1{#1}   \fi
\ifx \showURL      \undefined \def \showURL       #1{#1}          \fi
% The following commands are used for tagged output and should be
% invisible to TeX
\providecommand\bibfield[2]{#2}
\providecommand\bibinfo[2]{#2}
\providecommand\natexlab[1]{#1}
\providecommand\showeprint[2][]{arXiv:#2}

\bibitem[\protect\citeauthoryear{Auer}{Auer}{2003}]%
        {Auer2003}
\bibfield{author}{\bibinfo{person}{Peter Auer}.}
  \bibinfo{year}{2003}\natexlab{}.
\newblock \showarticletitle{{Using Confidence Bounds for
  Exploitation-Exploration Trade-offs}}.
\newblock \bibinfo{journal}{{\em Journal of Machine Learning Research\/}}
  \bibinfo{volume}{3} (\bibinfo{year}{2003}), \bibinfo{pages}{397--422}.
\newblock
\showISSN{15324435}


\bibitem[\protect\citeauthoryear{Behnke}{Behnke}{2003}]%
        {Behnke2003}
\bibfield{author}{\bibinfo{person}{Sven Behnke}.}
  \bibinfo{year}{2003}\natexlab{}.
\newblock \bibinfo{booktitle}{{\em {Hierarchical Neural Networks for Image
  Interpretation}}}.
\newblock
\showISBNx{9783540407225}


\bibitem[\protect\citeauthoryear{Breiman}{Breiman}{1996}]%
        {Breiman1996}
\bibfield{author}{\bibinfo{person}{Leo Breiman}.}
  \bibinfo{year}{1996}\natexlab{}.
\newblock \showarticletitle{{Bagging Predictors}}.
\newblock \bibinfo{journal}{{\em Machine Learning\/}} \bibinfo{volume}{24},
  \bibinfo{number}{421} (\bibinfo{year}{1996}), \bibinfo{pages}{123--140}.
\newblock
\showISSN{0885-6125}


\bibitem[\protect\citeauthoryear{Brezillon and Dwight}{Brezillon and
  Dwight}{2009}]%
        {brezillon2009aerodynamic}
\bibfield{author}{\bibinfo{person}{Jo{\"e}l Brezillon} {and}
  \bibinfo{person}{Richard~P. Dwight}.} \bibinfo{year}{2009}\natexlab{}.
\newblock \showarticletitle{Aerodynamic Shape Optimization Using the Discrete
  Adjoint of the Navier-Stokes Equations: Applications Toward Complex 3D
  Configutations}.
\newblock  (\bibinfo{year}{2009}).
\newblock


\bibitem[\protect\citeauthoryear{Brochu, Cora, and de~Freitas}{Brochu
  et~al\mbox{.}}{2010}]%
        {Brochu2010}
\bibfield{author}{\bibinfo{person}{Eric Brochu}, \bibinfo{person}{Vlad~M.
  Cora}, {and} \bibinfo{person}{Nando de Freitas}.}
  \bibinfo{year}{2010}\natexlab{}.
\newblock \showarticletitle{{A Tutorial on Bayesian Optimization of Expensive
  Cost Functions, with Application to Active User Modeling and Hierarchical
  Reinforcement Learning}}.
\newblock  (\bibinfo{year}{2010}).
\newblock


\bibitem[\protect\citeauthoryear{{\v{C}}repin{\v{s}}ek, Liu, and
  Mernik}{{\v{C}}repin{\v{s}}ek et~al\mbox{.}}{2013}]%
        {Crepinsek2013}
\bibfield{author}{\bibinfo{person}{Matej {\v{C}}repin{\v{s}}ek},
  \bibinfo{person}{Shih-Hsi Liu}, {and} \bibinfo{person}{Marjan Mernik}.}
  \bibinfo{year}{2013}\natexlab{}.
\newblock \showarticletitle{{Exploration and Exploitation in Evolutionary
  Algorithms: A Survey}}.
\newblock \bibinfo{journal}{{\it Comput. Surveys}} \bibinfo{volume}{45},
  \bibinfo{number}{3} (\bibinfo{year}{2013}), \bibinfo{pages}{35:1--35:33}.
\newblock
\showISBNx{03600300}
\showISSN{03600300}


\bibitem[\protect\citeauthoryear{Cully, Clune, Tarapore, and Mouret}{Cully
  et~al\mbox{.}}{2015}]%
        {Cully2015}
\bibfield{author}{\bibinfo{person}{Antoine Cully}, \bibinfo{person}{Jeff
  Clune}, \bibinfo{person}{Danesh Tarapore}, {and}
  \bibinfo{person}{Jean-Baptiste Mouret}.} \bibinfo{year}{2015}\natexlab{}.
\newblock \showarticletitle{Robots that can adapt like animals}.
\newblock \bibinfo{journal}{{\em Nature\/}} \bibinfo{volume}{521},
  \bibinfo{number}{7553} (\bibinfo{year}{2015}), \bibinfo{pages}{503--507}.
\newblock


\bibitem[\protect\citeauthoryear{Emmerich and Naujoks}{Emmerich and
  Naujoks}{2004}]%
        {Emmerich2004}
\bibfield{author}{\bibinfo{person}{Michael Emmerich} {and}
  \bibinfo{person}{Boris Naujoks}.} \bibinfo{year}{2004}\natexlab{}.
\newblock \showarticletitle{{Metamodel Assisted Multiobjective Optimisation
  Strategies and their Application in Airfoil Design}}.
\newblock \bibinfo{journal}{{\em Adaptive Computing in Design and Manufacture
  VI\/}} (\bibinfo{year}{2004}), \bibinfo{pages}{249--260}.
\newblock
\showISBNx{978-0-85729-338-1}


\bibitem[\protect\citeauthoryear{Gaier, Asteroth, and Mouret}{Gaier
  et~al\mbox{.}}{2017}]%
        {gaier2017feature}
\bibfield{author}{\bibinfo{person}{Adam Gaier}, \bibinfo{person}{Alexander
  Asteroth}, {and} \bibinfo{person}{Jean-Baptiste Mouret}.}
  \bibinfo{year}{2017}\natexlab{}.
\newblock \showarticletitle{Feature Space Modeling Through Surrogate
  Illumination}.
\newblock \bibinfo{journal}{{\em arXiv preprint arXiv:1702.03713\/}}
  (\bibinfo{year}{2017}).
\newblock


\bibitem[\protect\citeauthoryear{Gu}{Gu}{2016}]%
        {Gu2016}
\bibfield{author}{\bibinfo{person}{Shenkai Gu}.}
  \bibinfo{year}{2016}\natexlab{}.
\newblock \showarticletitle{{Multi-objective and Semi-supervised Heterogeneous
  Classifier Ensembles}}.
\newblock  \bibinfo{number}{April} (\bibinfo{year}{2016}).
\newblock


\bibitem[\protect\citeauthoryear{Hasenj{\"a}ger, Sendhoff, Sonoda, and
  Arima}{Hasenj{\"a}ger et~al\mbox{.}}{2005}]%
        {hasenjager2005three}
\bibfield{author}{\bibinfo{person}{Martina Hasenj{\"a}ger},
  \bibinfo{person}{Bernhard Sendhoff}, \bibinfo{person}{Toyotaka Sonoda}, {and}
  \bibinfo{person}{Toshiyuki Arima}.} \bibinfo{year}{2005}\natexlab{}.
\newblock \showarticletitle{Three dimensional evolutionary aerodynamic design
  optimization with CMA-ES}. In \bibinfo{booktitle}{{\em Proceedings of the 7th
  annual conference on Genetic and evolutionary computation}}. ACM,
  \bibinfo{pages}{2173--2180}.
\newblock


\bibitem[\protect\citeauthoryear{He, Zhang, Ren, and Sun}{He
  et~al\mbox{.}}{2016}]%
        {He2015}
\bibfield{author}{\bibinfo{person}{Kaiming He}, \bibinfo{person}{Xiangyu
  Zhang}, \bibinfo{person}{Shaoqing Ren}, {and} \bibinfo{person}{Jian Sun}.}
  \bibinfo{year}{2016}\natexlab{}.
\newblock \showarticletitle{Deep residual learning for image recognition}. In
  \bibinfo{booktitle}{{\em Proceedings of the IEEE Conference on Computer
  Vision and Pattern Recognition}}. \bibinfo{pages}{770--778}.
\newblock


\bibitem[\protect\citeauthoryear{Jin}{Jin}{2011}]%
        {Jin2011}
\bibfield{author}{\bibinfo{person}{Yaochu Jin}.}
  \bibinfo{year}{2011}\natexlab{}.
\newblock \showarticletitle{{Surrogate-assisted evolutionary computation:
  Recent advances and future challenges}}.
\newblock \bibinfo{journal}{{\em Swarm and Evolutionary Computation\/}}
  \bibinfo{volume}{1}, \bibinfo{number}{2} (\bibinfo{date}{jun}
  \bibinfo{year}{2011}), \bibinfo{pages}{61--70}.
\newblock
\showISSN{22106502}


\bibitem[\protect\citeauthoryear{Kamentsky and Liu}{Kamentsky and Liu}{1963}]%
        {kamentsky1963computer}
\bibfield{author}{\bibinfo{person}{Louis~A. Kamentsky} {and}
  \bibinfo{person}{Chao-Ning Liu}.} \bibinfo{year}{1963}\natexlab{}.
\newblock \showarticletitle{Computer-automated design of multifont print
  recognition logic}.
\newblock \bibinfo{journal}{{\em IBM Journal of Research and Development\/}}
  \bibinfo{volume}{7}, \bibinfo{number}{1} (\bibinfo{year}{1963}),
  \bibinfo{pages}{2--13}.
\newblock


\bibitem[\protect\citeauthoryear{Levenberg}{Levenberg}{1944}]%
        {levenberg1944method}
\bibfield{author}{\bibinfo{person}{Kenneth Levenberg}.}
  \bibinfo{year}{1944}\natexlab{}.
\newblock \showarticletitle{{A method for the solution of certain non--linear
  problems in least squares}}.
\newblock \bibinfo{journal}{{\it Quart. Appl. Math.}} \bibinfo{volume}{2},
  \bibinfo{number}{1} (\bibinfo{year}{1944}), \bibinfo{pages}{164--168}.
\newblock


\bibitem[\protect\citeauthoryear{Morse and Stanley}{Morse and Stanley}{2016}]%
        {Morse2016}
\bibfield{author}{\bibinfo{person}{Gregory Morse} {and}
  \bibinfo{person}{Kenneth~O Stanley}.} \bibinfo{year}{2016}\natexlab{}.
\newblock \showarticletitle{{Simple Evolutionary Optimization Can Rival
  Stochastic Gradient Descent in Neural Networks}}.
\newblock  \bibinfo{number}{Gecco} (\bibinfo{year}{2016}).
\newblock
\showISBNx{9781450342063}


\bibitem[\protect\citeauthoryear{Mouret and Clune}{Mouret and Clune}{2015}]%
        {Mouret2015}
\bibfield{author}{\bibinfo{person}{Jean-Baptiste Mouret} {and}
  \bibinfo{person}{Jeff Clune}.} \bibinfo{year}{2015}\natexlab{}.
\newblock \showarticletitle{Illuminating search spaces by mapping elites}.
\newblock \bibinfo{journal}{{\em arXiv preprint arXiv:1504.04909\/}}
  (\bibinfo{year}{2015}).
\newblock


\bibitem[\protect\citeauthoryear{Qui{\~n}onero-Candela and
  Rasmussen}{Qui{\~n}onero-Candela and Rasmussen}{2005}]%
        {Quinonero}
\bibfield{author}{\bibinfo{person}{Joaquin Qui{\~n}onero-Candela} {and}
  \bibinfo{person}{Carl~Edward Rasmussen}.} \bibinfo{year}{2005}\natexlab{}.
\newblock \showarticletitle{A unifying view of sparse approximate Gaussian
  process regression}.
\newblock \bibinfo{journal}{{\em Journal of Machine Learning Research\/}}
  \bibinfo{volume}{6}, \bibinfo{number}{Dec} (\bibinfo{year}{2005}),
  \bibinfo{pages}{1939--1959}.
\newblock


\bibitem[\protect\citeauthoryear{Rasmussen and Williams}{Rasmussen and
  Williams}{2006}]%
        {rasmussen2006gaussian}
\bibfield{author}{\bibinfo{person}{Carl Rasmussen} {and} \bibinfo{person}{Chris
  Williams}.} \bibinfo{year}{2006}\natexlab{}.
\newblock \showarticletitle{Gaussian Processes for Machine Learning}.
\newblock \bibinfo{journal}{{\em Gaussian Processes for Machine Learning\/}}
  (\bibinfo{year}{2006}).
\newblock


\bibitem[\protect\citeauthoryear{Snelson and Ghahramani}{Snelson and
  Ghahramani}{2006}]%
        {Snelson2006}
\bibfield{author}{\bibinfo{person}{Edward Snelson} {and}
  \bibinfo{person}{Zoubin Ghahramani}.} \bibinfo{year}{2006}\natexlab{}.
\newblock \showarticletitle{{Sparse Gaussian processes using pseudo-inputs}}.
\newblock \bibinfo{journal}{{\em Advances in Neural Information Processing
  Systems\/}} (\bibinfo{year}{2006}).
\newblock


\bibitem[\protect\citeauthoryear{{Yew-Soon Ong}, {Zongzhao Zhou}, and {Dudy
  Lim}}{{Yew-Soon Ong} et~al\mbox{.}}{2006}]%
        {Yew-SoonOng2006}
\bibfield{author}{\bibinfo{person}{{Yew-Soon Ong}}, \bibinfo{person}{{Zongzhao
  Zhou}}, {and} \bibinfo{person}{{Dudy Lim}}.} \bibinfo{year}{2006}\natexlab{}.
\newblock \showarticletitle{{Curse and Blessing of Uncertainty in Evolutionary
  Algorithm Using Approximation}}. In \bibinfo{booktitle}{{\em 2006 IEEE
  International Conference on Evolutionary Computation}}.
  \bibinfo{publisher}{IEEE}, \bibinfo{pages}{2928--2935}.
\newblock
\showISBNx{0-7803-9487-9}


\end{thebibliography}

\end{document}